\documentclass[12pt]{article}
\usepackage[english]{babel}
\usepackage{hyphenat}
\usepackage[linesnumbered,ruled,vlined]{algorithm2e}
\usepackage[utf8]{inputenc}
\usepackage{multirow,xcolor}
\usepackage{array, comment}
\usepackage{amssymb}
\usepackage{amsmath}
\usepackage{amsthm}
\theoremstyle{definition}

\usepackage{caption}
\usepackage{subcaption,graphicx}
\PassOptionsToPackage{hyphens}{url}
\PassOptionsToPackage{hyphens}{url}
\usepackage[colorlinks=true,citecolor=blue]{hyperref}
\usepackage{xurl}
\usepackage[margin=1in]{geometry}
\usepackage{tabularx}
\usepackage{setspace, enumitem, comment}
\usepackage{mathabx}
\usepackage{siunitx}
\usepackage{dsfont}
\usepackage{parskip}

\newcommand{\bs}{\boldsymbol}

\title{COVID-19  Sentiment Analysis \\ Using College Subreddit Data}
\author{Tian Yan, Fang Liu\\
\normalsize{Applied and Computational Mathematics and Statistics}\\
\normalsize{University of Notre Dame, Notre Dame, IN 46556}\\
\normalsize{fliu2@nd.edu}}
\date{ }

\begin{document}
\maketitle


\begin{abstract}
Background: The  COVID-19  pandemic  has  affected our society and human  well-being in  various  ways. In this study, we investigate how the pandemic has influenced people's emotions and psychological states compared to a pre-pandemic period using real-world data from social media. 

Method: We collected Reddit social media  data from 2019 (pre-pandemic) and 2020 (pandemic) from the subreddits communities associated with eight universities. We  applied the pre-trained  Robustly Optimized BERT pre-training approach (RoBERTa) to learn text embedding from the Reddit messages, and leveraged the relational information among posted messages to  train a graph attention network (GAT) for sentiment classification.  Finally, we applied model stacking to combine the prediction probabilities from RoBERTa and GAT to yield the final  classification on sentiment. With the model-predicted sentiment labels on the collected data, we used a generalized linear mixed-effects model to estimate the effects of pandemic and in-person teaching during the pandemic on sentiment. 

Results:  The results suggest that the odds of negative sentiments in 2020 (pandemic) were 25.7\% higher than the odds  in 2019 (pre-pandemic) with a $p$-value $<0.001$; and the odds of negative sentiments associated in-person learning were 48.3\% higher than with remote learning in 2020 with a $p$-value of 0.029.  

Conclusions: Our study results are consistent with the findings in the literature on the negative impacts of the pandemic on people's emotions and psychological states.  Our study contributes to the growing real-world evidence on the various negative impacts of the pandemic on our society; it also provides a good example of using both ML techniques and statistical modeling and inference to make  better use of real-world data.

\textbf{keywords}:  COVID, college,  graph neural networks, graph attention network, generalized linear
mixed-effects model, pandemic, RoBERTa, Reddit,  sentiment
\end{abstract}

\section{Introduction}
The COVID-19 pandemic has affected our society in many ways -- traveling was limited and restricted, supply chains were disrupted, companies experienced contractions in production,  financial markets were not stable, school was temporarily closed, and in-person learning was replaced by remote and online learning. Last but not the least, the pandemic has taken a toll on our physical health and also has had a huge impact on mention health, found by multiple studies. \cite{xiong2020impact} provided a systematic review on how the pandemic has led to increased symptoms of anxiety, depression, post-traumatic stress disorder in the general population.  \cite{bo2021posttraumatic} observed that COVID-19 patients suffered from post-traumatic stress symptoms before discharge. \cite{zhang2020differential} detected an increased prevalence of depression  predominately in COVID-19 patients. \cite{chen2020prevalence} observed that the prevalence of self-reported depression and anxiety among pediatric medical staff members was significantly high during the pandemic, especially for workers who had COVID-19 exposure experience. Using online surveys, \cite{sonderskov2020depressive} claimed that the psychological well-being of the general Danish population is negatively  affected by the COVID-19 pandemic.  \cite{wu2020analysis}  surveyed  college students from the top 20 universities in China and found the main causes of anxiety among college students included online learning and epidemic diseases.  \cite{sharma2020assessing} collected text data via an APP and  learned that there was a high rate of negative sentiments among students and they were  more interested in health-related  topics and  less interested in education during the pandemic. Sentiment analysis using social media data is a commonly used approach with a wide range of applications such as epidemic detection, infection break localization, medical health informatics system development, and vaccine hesitancy identification and invention, among others. Readers may refer to \cite{alamoodi2021multi} and \cite{alamoodi2021sentiment} for reviews on the this topic. 

The above listed work uses surveys, patient data from hospitals and health care providers, or data collected through  specifically designed apps to study the impact of the pandemic on the mental states in various demographic groups. Social media data, on the other hand, provide another great source of real-world data and contain a vast amount of information that can be leveraged to study the effects of the pandemic on people's life, behaviors, and health, among others. Due to the unstructured nature of the text and semantic data collected from social media, text mining and machine learning (ML) techniques in natural language processing (NLP) are often used to make sense of the data.  For example,  \cite{low2020natural} applied  sentiment analysis, classification and clustering, topic modeling to uncover concerns throughout Reddit before and during the pandemic.   \cite{jelodar2020deep} applied long short-term memory (LSTM) recurrent neural networks (RNNs) to Reddit data and achieved an 81.15\%  sentiment classification accuracy on COVID-19 related comments. \cite{jia2020emotional}  collected and analyzed data from Weibo (a Chinese social media platform) and concluded that students' attitude toward returning to school was positive.  \cite{pandey2021redbert} applied the Bidirectional Encoder Representations from Transformers (BERT) to Reddit data to assimilate meaningful latent topics on COVID-19 and sentiment classification.

Besides the rich text and semantic information in social media data, the data are also known for their vast amount of relational information that is also important for training effective ML procedures.  Relational information is often formulated as networks or graphs.  Graph neural networks (GNNs), a type of NNs that take graphs as input for various learning tasks such as node classification and graph embedding, can be used for learning from  both the semantic and relational information in social media data. Since \cite{scarselli2008graph} proposed the first GNN,  many GNN extensions and variants have been proposed.  \cite{li2015gated} incorporated gated recurrent units into GNNs.  \cite{defferrard2016convolutional} generalized convolutional NNs (CNNs) to graphs using spectral graph theory.  \cite{kipf2016semi} proposed a semi-supervised layer-wise spectral model to effectively utilize graph structures. \cite{atwood2016diffusion} introduced the diffusion-convolution operation to CNNs to learn representation of graphs. \cite{monti2017geometric} generalized CNNs to non-Euclidean structured data including graphs and manifolds. \cite{hamilton2017inductive} extended semi-supervised learning to a general spatial CNN by sampling from local neighborhoods, aggregating features, and exploring different aggregation functions.  \cite{velivckovic2017graph} proposed graph attention networks (GAT), a self-attention-based GNN to assign different weights to different neighbors. It is computationally efficient because of parallelization and doesn't require knowing the overall graph structure upfront, but it works with homogeneous graphs only.  \cite{wang2019heterogeneous} developed the heterogeneous attention network (HAN) that extends GAT to heterogeneous graphs of  multi-type nodes and edges. They employ the attention mechanisms to aggregate different types of neighbors on different types of meta-paths that represent different types of relations among nodes of different or the same types. HAN is capable of paying attention to ``important'' neighbors for a given node and capturing meaningful semantics efficiently. \cite{hu2020heterogeneous} extended the HAN framework to learn from dynamic heterogeneous graphs. \cite{chen2020simple} improved previous GNN approaches by using initial residual connections and identity mapping and mitigating over-smoothing.

In this work, we leverage the advances in NLP and GNNs to learn from social media data and study whether the pandemic has negatively affected the emotions and psychological states of people who are connected with a higher-education  institute (HEI). ``Connected with an HEI'' in this context is defined in a very broad sense, described as  follows. The data we collected  are from university subreddits communities that include basically everyone who contributes to the subreddits during August to November 2019 and August to November 2020 (August to November is when most schools in  the US are in session). Therefore, an individual who contributes to our data might be a student in an HEI, a staff or faculty member there, or people who do have a direct connection with the HEI but  is interested in the HEI or are involved in some situations some way. It is not our goal to generalize the conclusion from this study to the general population as the studied population is clearly not a representative sample of the whole population. However, it represents a subgroup of the general population to which the study conclusions can be generalized to and contributes to the body of literature  and research on the impact of the pandemic on mental health in various sub-populations using real-world data. 

We collected the social media data in 2019 (pre-pandemic) and 2020 (pandemic) from the subreddits communities associated with 8 universities chosen per a full-factorial design according to three factors as detailed in Section \ref{sec:datacollection}. 
We employ the Robustly Optimized BERT pre-training approach (RoBERTa) and GNNs with the attention mechanism that takes both semantic and relational information as input to perform sentiment analysis in a semi-supervised manner to save manual labeling costs. The predicted sentiment labels are then analyzed by a generalized linear mixed model to examine the effect of pandemic and online learning on the mental state of the communities represented by the collected data. Our contributions and the main results are summarized as follows.

\begin{itemize}[leftmargin=18pt]
\setlength{\itemsep}{-3pt}
\item We adopt a full-factorial $2\times2\times2$ design when choosing schools for Reddit data collection. This not only makes the collected data more representative by covering different types of schools compared to using data from a single HEI,  but also helps with controlling for confounders  for the subsequent statistical inference on the effect of the pandemic on the outcome of interest.
\item We use both the semantic and graph information from the collected Reddit data as inputs, leverage the state-of-the-art NLP techniques and GNNs with the attention mechanism, and apply model stacking to further improve the sentiment classification accuracy. 
\item Our results suggest that the odds of negative sentiments increased by 25.7\% during the pandemic compared to the pre-pandemic period in a statistically significant manner ($p$-value $<0.001$ from comparing the sentiments of 91,200 cases in 2020 vs 74,370 cases in 2019). During the pandemic, the odds of  negative sentiments in the schools that opted for in-person learning were 48.3\%  statistically significantly higher than that in the schools that chose remote learning ($p$-value = 0.029 from comparing the sentiments of 49,512 cases for in-person learning vs 41,688  cases for remote learning in 2020). 
\end{itemize}

The rest of the paper is organized as follows. Section \ref{sec:method} describes the research method, including data collection and pre-processing, the ML procedures for sentiment classification, and the model for hypothesis testing and statistical inference on the effects of the pandemic on sentiment. The main results are presented in   Section \ref{sec:results}. Discussions of the study including its limitation and future directions are provided in Section \ref{sec:discussion}. The conclusions and some final remarks are presented in  Section \ref{sec:conlusions}.

\section{Methods}\label{sec:method}
Fig 1 depicts the process and main steps of our  research method in this study.  We feed the collected text data from Reddit to a pre-trained RoBERTa, which produces embeddings that contain the meaning of each word in the input text  data. The embeddings are used for two downstream tasks. First, they are sent to a classification layer (softmax function) to predict the  probabilities of negative and non-negative sentiments; second, they are used as part of the input, along with adjacency matrices formulated from the relational data among the messages collected from Reddit, to a GNN. The GNN outputs another set of predicted  probabilities of negative and non-negative sentiments, which are combined with the set of probabilities from RoBERTa to train a meta-model (the ensemble and model stacking step) to produce a final classifier. The classifier is then used to generate a sentiment label for each unlabeled message in our collected data. Finally, we combine the data with the observed or learned sentiment labels across the 8 schools prior to and during the pandemic, and employ regression (generalized linear mixed-effects model or GLMM) to examine the effect of the pandemic on sentiment, after adjusting for school characteristics. 

In what follows, we illustrate each step in details.  Section \ref{sec:datacollection} and Section \ref{sec:labeling} present the data collection and  manual labelling steps to create a set of labeled data to train the ML procedures, respectively;  Section \ref{sec:RoBERTa}, Section \ref{sec:GAT}, and Section \ref{sec:ensemble} present the application of RoBERTa, formulation and training of GAT, and model stacking and ensemble of RoBERTa and GAT, respectively;  Section \ref{sec:inference} lists the GLMM used for statistical testing and inference.

\subsection{Data Collection}\label{sec:datacollection}
Our data collection and how the data are used in this study are in accordance with Reddit’s Terms and Conditions on data collection and usage. We also consulted the research compliance program at the University of Notre Dame and no IRB approval is needed given that the collected data are publicly accessible on Reddit; we did not collect private identifiable data nor interacted with the Reddit users. More information regarding the privacy compliance is provided in Appendix \ref{S1_File}.

In the current study, we focus on schools that are ``R1: Doctoral Universities – Very high research activity" per the Carnegie Classification of Institutions of Higher Education  \cite{CC} in this study, with the plan to examine more universities in the future. A university can be  characterized in many different ways. When choosing which schools to include in our study, we focus on three factors that can potentially affect students' sentiment:  private vs.~public schools, school location (small vs.~large cities), and whether a school opted in for in-person learning during the pandemic vs.~taking a fully online learning approach from August to November in 2020. To balance potential confounders for the subsequent statistical analysis, we adopt a full-factorial $2\times2\times2$ design on the above three factors  and select one school for each of the eight cells in the full-factorial design (Table \ref{tab:Reddits}). Each of the eight schools has a subreddit on Reddit.   We downloaded messages, including original posts and comments, post replies, and comment replies, from each subreddit from August to November in 2019 and 2020, respectively, representing the pre-pandemic and pandemic periods, using the Pushshift API (\url{https://github.com/pushshift/api}). This results in 165,570 messages in total. 

\begin{table}[!ht]
\begin{center}
\caption{\bf Characteristics of the 8 schools in this study } \label{tab:Reddits}\vspace{-3pt}
\resizebox{1.0\textwidth}{!}{
\begin{tabular}{@{\hspace{5pt}}l|c@{\hspace{5pt}}c@{\hspace{5pt}}c@{\hspace{5pt}}c@{\hspace{6pt}}c@{\hspace{5pt}}c@{\hspace{5pt}}c@{\hspace{3pt}}c@{\hspace{2pt}}} 
\hline
School & UCLA & UCSD & UCB &  UMich  & Harvard & Columbia & Dartmouth & ND\\
\hline
funding & public &public &public &public &private  &private  &private  &private \\
\hline
location & large &large &small &small &large & large &small &small\\
\hline
in-person learning & no & yes & no &yes & no& yes& no& yes\\
Aug$\sim$Nov, 2020\\
\hline
\# of messages 2019 & 16,894 & 17,502 & 16,673 & 12,524 & 2,893 & 5,031 & 1,274 & 1,579 \\
\cline{2-9}
\# of messages 2020 & 18,714 & 17,346 & 18,194 & 14,879 & 3,554 & 9,337 & 1,226 & 7,950 \\
\hline
Data period & \multicolumn{8}{c}{Aug to Nov 2019; Aug to Nov 2020}\\
\hline
\end{tabular}}
\resizebox{1.0\textwidth}{!}{
\begin{tabular}{l}
UCB = UC-Berkley; UMich = U of Michigan; ND= Notre Dame \textcolor{white}{.\hspace{3in}.}\\
\hline
\end{tabular}}
\end{center}
\end{table}

\subsection{Manual Labeling}\label{sec:labeling}
For the downloaded data, we manually labeled a sizable subset of the messages in each school-year with sufficient labeled cases in each sentiment category for training and testing of our ML procedures.  The labeled messages are summarized in Table \ref{tab:label}. In term of the labelling effort per se, we recruited 7 volunteer labellers. The labellers were given a written guidance and a set of rules to guide their labelling effort. We accessed the agreement rate between several pairs of labellers. There was a 84.04\% overall agreement rate with a Cohen's kappa coefficient of 0.634.  Both the agreement rates and the kappa coefficient suggest good agreement among the raters. 
\begin{table}[!ht]
\centering 
\caption{\bf Manually labeled sentiment category frequencies in each school-year}\label{tab:label}
\vspace{-3pt}
\resizebox{0.625\textwidth}{!}{
\begin{tabular}{lcccc} 
\hline
School &Year& non-Negative & Negative & Total \\
\hline
\multirow{2}{5em}{UCLA}&2019& 178 & 53 & 231 \\
& 2020 & 170 &52 & 222 \\
\hline
\multirow{2}{5em}{UCSD}&2019& 145 &50 &195 \\
& 2020 & 130 & 66&196 \\
\hline
\multirow{2}{5em}{UCB}&2019& 146 & 63 & 209 \\
& 2020 & 161 &52 & 213 \\
\hline
\multirow{2}{5em}{UMich}&2019& 134  &80 &214 \\
& 2020 & 117 &82 &199 \\
\hline
\multirow{2}{5em}{Harvard}&2019& 164 & 50&214 \\
& 2020 & 150 & 58 & 208 \\
\hline
\multirow{2}{5em}{Columbia}&2019& 127 & 50&177 \\
& 2020 & 118 &54 &172 \\
\hline
\multirow{2}{5em}{Dartmouth}&2019& 324 & 78 & 402 \\
& 2020 & 327 &66 &393 \\
\hline
\multirow{2}{5em}{ND}&2019& 187 &52 &239 \\
& 2020 & 124 &103 & 227 \\
\hline
\multirow{2}{5em}{total} & 2019 & 1405 & 476 &  1881 \\
  & 2020 & 1297 & 533 & 1830 \\
\hline
\end{tabular}}
\end{table}

We focus on binary sentiment classification in this study. Instead of the  Negative vs Positive classification, we use the Negative vs non-Negative classification because some messages are rather neutral than being positive or negative. For example, the message could just be a question on the number of students in a class, the reply to which is a number, often implying no negative or positive sentiment. In addition, there is often subjectivity among manual labelers on what Neural means. For example, some labelers label ``O.K.'' as Positive but others labeled it as Neutral. We had also examined the possibility of adding the Very Positive and Very Negative categories, and there was substantial disagreement among the labelers on what can be described as Very Positive vs Positive and Very Negative vs Negative. We also examined the classification accuracy on sentiments when using 3 categories (Positive, Negative, Neutral) vs using 5 categories  (Very Positive, Positive, Negative, Very  Negative, Neutral), the prediction accuracy was  $\sim60\%$ for the former and $\sim50\%$ for the latter. Our results are consistent with the findings in the literature; that is, $45\%\sim55\%$ for in  the 5-category classification \cite{socher2013recursive,mccann2017learned}.
Given the objective of our study -- whether the pandemic has negatively affected the emotions and psychological states of people -- and that the 2-class classification  is commonly accepted in sentiment analysis, we expect that the Negative versus non-Negative classification is sufficient to address our study goal, along with less discrepancy when labelling and higher accuracy when training the ML procedures and predicting the messages in our collected data.   

\subsection{Text Embedding via RoBERTa}\label{sec:RoBERTa}
RoBERTa  \cite{liu2019roberta}  is an improved version of the BERT (Bidirectional Encoder Representations from Transformers) framework  \cite{devlin2018bert}, owing to several modifications such as dynamic masking, input changes without the next-sentence-prediction loss, large mini-batches, etc. The BERT framework itself is based on transformers  \cite{vaswani2017attention}, a deep learning model built upon a number of encoder layers and multi-heads self-attention mechanism.  At the end of the training process, BERT learns contextual embeddings for words. The BERT model is pre-trained using unlabeled data extracted from the BooksCorpus (800 million words) and English Wikipedia (2,500 million words) and can be fine-tuned for various types of downstream learning tasks.  

The Reddit data in our study, the same as any social media data, contain a large number of emoticons, non-standard spellings and internet slang, causing difficulty for traditional sentence embedding learning methods designed for semantics with standard grammar and spelling. The RoBERTa model that we employ is capable of generating embedding from internet slang,  and other non-standard spelling  by using special tokenizer.  Specifically, we applied the RoBERTa model trained on $\sim\!\!58$ million messages from Twitter and fine-tuned for sentiment analysis   \cite{barbieri2020tweeteval, modelcodes} to obtain the embeddings of the semantic information in our collected Reddit text data. 

Besides generating embeddings from the text data via RoBERTa, which are fed to the GAT in the next step, we also use the RoBERTa framework to  predict the labels of the sentiments in the collected Reddit data, as part of the model stacking step introduced in  Section \ref{sec:ensemble}. The  Python  code  for  the  RoBERTa framework that we applied  is  adapted  from \cite{barbieri2020tweeteval} and is available at \url{https://huggingface.co/cardiffnlp/twitter-roberta-base-sentiment}

\subsection{Training of GAT}\label{sec:GAT}
We employ GAT to incorporate the relational information in the collected data as a graph input to train a sentiment classifier.  GAT is a GNN that employs the attention mechanism to incorporate neighbors' information in graph embedding.  In our study, each message is regarded as a node in the graph and the relation of ``message $\xrightarrow{\text{reply to}}$ message'' is coded as an edge in the graph and 1 in the corresponding adjacency matrix (we tried both asymmetric and symmetric adjacency matrices as input data when training the GAT, the former does not yield better results; we reported the results from coding the relations as a symmetric adjacency matrix).  Neighbors of a node are those who are connected with the node with an edge. 

We randomly selected 601 non-Negative and 94 Negative messages from the combined 2019 and 2020 labelled Dartmouth data in Table \ref{tab:label}, along with the adjacency matrix among the collected comments from Dartmouth (whether labelled nor not), as the training set for GAT.  There are a couple of reasons why we used the data from one school to train GAT. First, the graph of the merged data across different schools leads to a high-dimensional block-diagonal adjacency matrix because the comments from different schools are barely linked. In other words, merging the data from different schools does not provide additional benefits from the perspective of leveraging relational data to train GAT, but potentially increasing computational costs, and, in our experiments, also leading to worse prediction accuracy than using the data from a single school. Second, laborious manual labeling confined us to getting sufficient labeled data  in only one school to train the GAT with meaningful relational information. Third, what makes negative vs positive sentiments and how they relate the relational information among the message is rather independent of a particular school; and there would not be much scarification in prediction accuracy via a trained GAT NN with the relational data from one school.  

The  steps of the training of the GAT NN are listed below and a schematic illustration of the steps is given in Fig 2.  There are two types of inputs to GAT; one is the embeddings $\mathbf{h}_i$ (a column vector) in message  $i$, learned via RoBERTa in the previous step; the other is the graph formed from the relations among the messages.  The output is improved embedding $\mathbf{z}_i$. The most important component of the GAT NN is the multi-head attention mechanism, applied to further improve the embedding $\mathbf{h}_i$ by incorporating the edge information in the graph. The implementation details are given below. 
\begin{enumerate}[leftmargin=18pt]
\setlength{\itemsep}{0pt}
\item[1)] Let $\mathcal{N}_i$ denote the neighbors of message $i$, that is, the set of messages that reply to message $i$. Define the \emph{importance} of message $j\in\mathcal{N}_i$ to message  $i$ as  
\begin{equation}\label{eqn:eij}
e^k_{ij}\triangleq\mathbf{a}^k\left(\mathbf{M}^k\mathbf{h}_i,\mathbf{M}^k\mathbf{h}_j\right).
\end{equation}
$\mathbf{M}^k$ is a linear transformation matrix (unknown and to be estimated) for $k=1,\ldots,K$, the index of ``heads'' in the multi-head attention mechanism. Using a multi-head mechanism (i.e., $K>1$) would help explore different sub-spaces of the input embedding for more robustness and stability compared to using a single head attention mechanism (i.e., $K=1$);  we use $K=4$ in our study.   $\mathbf{a}^k$ is the attention mechanism; we employ a single-layer feedforward NN parametrized by a weight vector (unknown and to be estimated) with the LeakyReLU  activation function and input $[\mathbf{M}^k\mathbf{h}_i\|\mathbf{M}^k\mathbf{h}_j]$, where
$\|$ denotes the row concatenation. The output $e_{ij}$ quantifies the importance of node $j$ to node $i$ in the graph and asymmetric (that is, $e_{ij}\neq e_{ji}$, the latter measures the importance of nodes $i$ to $j$).  To make $e_{ij}$ more comparable across the nodes $j\in\mathcal{N}_i$, we normalize $e_{ij}$  using the softmax function 
\begin{align}
\alpha^k_{ij}&=\frac{\exp(e^k_{ij})}{\sum_{j'\in\mathcal{N}_i} \exp(e^k_{ij'})}\mbox{ for each $j\!\in\!\mathcal{N}_i$}. \label{eqn:alphaij}
\end{align}
In summary, this step uses the embedding from the RoBERTa and the relations in the graph to determine the relative importance of one node to another. 

\item[2)] Compute a weighted linear average of the transformed input embeddings for all nodes $j\in\mathcal{N}_i$ in each head and then concatenate the average across the $K$ heads to generate the final embedding for node $i$; that is,
\begin{align}\label{eqn:zi}
\mathbf{z}_i&= \|^K_{k=1} \sigma\!\left(\textstyle \sum_{j\in \mathcal{N}_i}\!\alpha_{ij} \mathbf{M}^k\mathbf{h}_j\right), 
\end{align}
where the weights are $\alpha^k_{ij}$ from Eq \eqref{eqn:alphaij} and $\sigma$ is an activation function. 
\item[3)] Feed the output embedding $\mathbf{z}_i$ from step 2) for $i\!=\!1,\ldots,|\mathcal{Y}_{\text{L}}|$, where the set $\mathcal{Y}_{\text{L}}$ contains the labeled messages and $\mathcal{Y}_{\text{L}}=\{\mathcal{Y}_{\text{L}}^{\text{non-neg}},\mathcal{Y}_{\text{L}}^{\text{neg}}\}$ through a dense single layer perceptron $f$, parameterized by $\bs{\theta}_f$, and train $f$ by minimizing the $l_2$-regularized cross-entropy loss in Eq (\ref{eq:loss_aug}),
\begin{align}
L&\!=\!-\big(|\mathcal{Y}_{\text{L}}^{\text{non-neg}}|\!+\!r|\mathcal{Y}_{\text{L}}^{\text{neg}}|\big)^{-1}\!\bigg(\!
\sum_{i\in\mathcal{Y}_{\text{L}}^{\text{non-neg}}}\!\!\!\!\!\!\!y_i\!\log\!\big(f(\mathbf{z}_i(\bs\theta_z),\bs{\theta}_f\!)\!\big)\!+\!
r\!\!\!\!\sum_{i\in\mathcal{Y}_{\text{L}}^{\text{neg}}}\!\!\!y_i\log\!\big(f(\mathbf{z}_i(\bs\theta_z),\bs{\theta}_f\!)\!\big)\!\!\bigg)\notag\\
&\quad+\!\|\Theta\|^2_2.\label{eq:loss_aug}
\end{align}
$y_i$ denotes the observed label of message $i$ in the training data. $f$ classifies the messages into either Negative or non-Negative and is parameterized by $\bs\theta_f$, $\bs\theta_z$ contains the unknown parameters from the graph embedding and attention mechanism in  steps 1) and 2) above, and $\Theta={(\bs{\theta}_f,\bs{\theta}_z)^T}$.  $r$ is the weight parameter and is set at 1 in a general case. Our training data are imbalanced in terms of labeled non-Negative cases vs Negative cases (the former is 6 folds higher). To achieve better classification results, we ``oversampled'' the Negative cases, setting $r=2$. The $l_2$ penalty is $\lambda=0.08$, chosen via the 4-fold cross-validation.  The training was performed in TensorFlow with randomly initialized model parameters and the Adam optimizer  \cite{kingma2014adam} with a learning rate of 0.02. 
\end{enumerate}

The loss function in  Eq (\ref{eq:loss_aug}) is formulated with the labeled data only. Though some nodes in the input graph are unlabeled (given the high cost of manual labeling) and are not part of the loss function, their text information and their relational information with the labeled nodes are still used in the graph embedding in steps 1) and 2) above. In this sense, the training of the GAT NN can be regarded as semi-supervised learning.  The Python code for training the GAT NN is adapted from \cite{wang2019heterogeneous} (\url{https://github.com/Jhy1993/HAN}).

\subsection{Performance of GAT and RoBERTa and Model Stacking}\label{sec:ensemble}
The performance of the trained GAT and RoBERTa in sentiment predictions are examined and compared on the testing data set with 375 Negative and 375 non-Negative cases that are made of 25  randomly chosen  Negative cases and 25  randomly chosen  non-Negative cases from the combined 2019/2020 Dartmouth dataset that are not part of the training data and each of the rest 14 school-year datasets, respectively. For GAT, the  adjacency matrix of the 750 cases in the testing data is also part of the testing data.

Comparing the classification results by GAT and  RoBERTa on the testing data side by side,  we found  some inconsistency, which motivated us to develop a model stacking approach  to create an ensemble of RoBERTa and GAT so as to achieve more robust sentiment prediction. Model stacking \cite{wolpert1992stacked}  is an ensemble learning technique (ensemble learning combines multiple base models or weaker learners to form a strong learner; bagging, boosting, and model stacking are all well-known and popular ensemble methods and concepts). We first show the inconsistent results between RoBERTa and GAT and then present the model stacking approach.

The left plot  in Fig 3  shows the sensitivity or the probability of correctly predicting non-Negative sentiments; the right plot depicts the specificity, which is the probability of correctly predicting negative sentiments. The cutoff $c$ for labeling a message Negative or non-Negative based on $p$, the predicted probability of it being non-Negative, is optimized by  maximizing the geometric mean of sensitivity (recall) and positive predictive value (precision)  \cite{fowlkes1983method} in a ROC curve. $c$ is 0.4283 and 0.6923 for GAT and RoBERTa, respectively, in our study. If $p<c$, the message is labeled Negative; otherwise, it is labeled non-Negative.  In both plots, there is some deviation from the identity line, indicating the inconsistency in the classification between GAT and RoBERTa. GAT performs better  that RoBERTa in terms of predicting true negative sentiments, whereas RoBERTa performs better that GAT in terms of  predicting true non-negative sentiments. We also examined a set of messages where the discordance in label prediction between RoBERTa and GAT occurred and no obvious patterns were found.  In addition, both classifiers output some low probabilities; ideally, if both classifiers had high accuracy, most of the points would be clustered in the upper right corner around $(1,1)$. 

The observations in Fig 3  suggest that both GAT and RoBERTa have their respective strength but also suffer some degree of inaccuracy when predicting a certain sentiment category. To leverage the advantage of both methods and for more robust predictions, we develop a model stacking  method  by fitting  logistic regression models, a.k.a. meta-models, on the predicted classification probabilities from GAT and RoBERTa.  The steps of the model stacking application are listed below.  

We first scale the raw prediction probability $p$ of being non-Negative from GAT and RoBERTa to obtain $p'$.  Specifically, 
\begin{equation}
p'^{(m)}_{ijk}=0.5\times\frac{p^{(m)}_{ijk}-c^{(m)}}{1-c^{(m)}}+0.5,
\end{equation}
where $m=1,2$ represents the RoBERTa and GAT, respectively, $p^{(m)}_{ijk}$ is the predicted probability that message $i$ from school $j$ in year $k$ has a non-negative sentiment per the prediction by $m$,  The reason behind the scaling of $p$ is that the optimal $c^{(m)}$ for the raw $p$ is different for RoBERTa and GAT and neither equals 0.5,  leading to ineffective downstream training of the logistic regression meta-model (Eq (\ref{eqn:meta})) and difficulty in choosing a cutoff on the final probability $\bar{p}$ from the meta-model. After the scaling, the cutoff on $p'$ is 0.5 for both GAT and RoBERT, eliminating both concerns. 

Specifically, the logistic meta-model for school $j$ and year $k$ uses  $p'^{(1)}_{ijk}$ and $p'^{(2)}_{ijk}$ as input to generate the final prediction probability 
\begin{equation}\label{eqn:meta}
\bar{p}_{ijk}=\frac
{\exp\big(\hat{\beta}_{0}+\hat{\beta}_{1}p'^{(1)}_{ijk}+ \hat{\beta}_{2}p'^{(2)}_{ijk}\big)}
{1+\exp\big(\hat{\beta}_{0}+\hat{\beta}_{1}p'^{(1)}_{ijk}+\hat{\beta}_{2}p'^{(2)}_{ijk}\big)},
\end{equation}
where $\hat{\beta}_{0}$, $\hat{\beta}_{1}$, $\hat{\beta}_{2}$  are the estimated regression coefficients from the logistic model for all schools and years. The cutoff on $\bar{p}$ from the meta-model prediction is 0.5; that is, if $\bar{p}_{ijk}\ge0.5$, then message $i$ in school $j$ and year $k$ is non-Negative; otherwise, it is labeled as Negative.  

The testing data set for the logistic meta-model remains the same as the testing data for RoBERTa and GAT. The training data for model stacking include the remaining 25 Negative cases and 25 non-Negative cases  from the combined 2019/2029 Dartmouth data that are not part of the training data for GAT or the testing data and 25 Negative cases and 25 non-Negative cases from each of the rest 14 school-year data subsets that are not part of the testing data, respectively, leading to a total of 375 Negative and 375 non-Negative cases overall. 

\subsection{Statistical Modelling and Inference}\label{sec:inference}
With the learned sentiment labels (Negative vs. non-Negative) for the messages in the data set, we apply GLMM to examine the effects of the pandemic on emotional state. GLMM uses a logit link function with a binary outcome (non-Negative vs. Negative). Specifically, we run two GLMMs. Model 1 is fitted to the whole collected data of  165,570 cases and compares the sentiment between 2019 (pre-pandemic) and 2020 (pandemic), after controlling for school location and type. Model 2 is fitted to the 2020 subset data with 91,200 cases and examines how in-person learning during the pandemic affects the sentiment compared to remote learning, after controlling for school location and type. Since the messages are clustered by school and the messages from the same school are not independent, we include in both models a random effect of school, thus the ``mixed-effects'' model. The formulations of the two models are given below.
\begin{align}
\text{log}\left({\frac{\Pr(\text{non-Negative})}{1-\Pr(\text{non-Negative})}}\right)&=\!\beta_0\!+\!\beta_1 X_{\text{type}}\!+\!\beta_2 X_{\text{location}}\!+\!\beta_3 X_{\text{year}}\!+\!\gamma Z_{\text{school}},\\
\text{log}\left({\frac{\Pr(\text{non-Negative})}{1-\Pr(\text{non-Negative})}}\right)&=\!\beta'_0\!+\!\beta'_1 X_{\text{type}}\!+\!\beta'_2 X_{\text{location}}\!+\!\beta'_3 X_{\text{in-person}}\!+\!\gamma' Z_{\text{school}},\!
\end{align}
where $\gamma\sim\mathcal{N}(0,\sigma^2_{\gamma})$ and $\gamma'\sim\mathcal{N}(0,\sigma^2_{\gamma'})$, $X_{\text{type}}=1$ if private and 0 if public, $X_{\text{location}}=1$ if small city and 0 if large city, $X_{\text{year}}=1$ if 2020 and 0 if 2019, $X_{\text{in-person}}=1$ if it is in-person and 0 if it is remote, and $\beta_j$ and $\beta'_j$ are the corresponding fixed-effects for $j=1,2,3$ and represents the log odds ratio of being non-Negative when $X_j=1$  vs. $X_j=0$ in the models.

\section{Results}\label{sec:results}
\subsection{Sentiment classification}
The classification results from the meta-model from the model stacking on the testing data set are presented in Table \ref{tab:all} and Fig 4  with the results from RoBERTa and GAT as a comparison. The information conveyed by Fig 4 and Table \ref{tab:all} is about the same but with different presentations to compare the 3 classification approaches.  We examine multiple metrics on the  prediction accuracy, which include the overall accuracy rate, the F1 score, the positive predictive value (PPV or recall),  sensitivity (or precision), and specificity. 
\begin{table}[!ht]
\centering   
\caption{\bf Prediction accuracy on testing data}\label{tab:all}
\resizebox{0.95\linewidth}{!}{
\begin{tabular}{@{}c@{\hspace{6pt}}ccccc@{\hspace{4pt}} cc@{\hspace{6pt}}c@{\hspace{3pt}}c@{\hspace{6pt}} c@{}}
\hline
Metric & Year & \multicolumn{8}{c}{School}\\
 \cline{3-10}
& & UCLA &UCSD & UCB & UMich & Harvard & Columbia & Dartmouth\textsuperscript{\#} & ND & overall\\
\cline{3-10}
 & & \multicolumn{9}{c}{\textbf{meta-model via model stacking (ensemble)}}\\
 \cline{3-11}\hline
CAR\textsuperscript{*} & 2019 & 0.880& 0.840& 0.760& 0.740& 0.940& 0.800& 0.820& 0.820& 0.827\\
    & 2020 & 0.780& 0.820& 0.780& 0.840& 0.800& 0.920& & 0.860& \\
F1\textsuperscript{$\dagger$} & 2019 & 0.870& 0.818& 0.778& 0.735& 0.939& 0.792& 0.824& 0.816& 0.824\\
    & 2020 & 0.776& 0.824& 0.756& 0.846& 0.808& 0.920& & 0.863& \\
precision\textsuperscript{$\ddagger$} & 2019 & 0.952& 0.947& 0.724& 0.750& 0.958& 0.826& 0.808& 0.833& 0.836\\
    & 2020 & 0.792& 0.808& 0.850& 0.815& 0.778& 0.920& & 0.846& \\
recall\textsuperscript{${\dagger\dagger}$} & 2019 & 0.800& 0.720& 0.840& 0.720& 0.920& 0.760& 0.840& 0.800& 0.813\\
    & 2020 & 0.760& 0.840& 0.680& 0.880& 0.840& 0.920& & 0.880& \\
specificity & 2019 & 0.960& 0.960& 0.680& 0.760& 0.960& 0.840& 0.800& 0.840& 0.840\\
    & 2020 & 0.800& 0.800& 0.880& 0.800& 0.760& 0.920& & 0.840& \\
\hline
 & & \multicolumn{9}{c}{\textbf{RoBerta}}\\
\hline
CAR\textsuperscript{*} & 2019 & 0.860& 0.840& 0.780& 0.740& 0.940& 0.800& 0.780& 0.780& 0.823\\
    & 2020 & 0.820& 0.840& 0.800& 0.840& 0.780& 0.920& & 0.820& \\
F1\textsuperscript{$\dagger$} & 2019 & 0.851& 0.810& 0.800& 0.735& 0.939& 0.800& 0.800& 0.784& 0.823\\
    & 2020 & 0.830& 0.840& 0.783& 0.846& 0.784& 0.920& & 0.830& \\
precision\textsuperscript{$\ddagger$} & 2019 & 0.909& 1.000& 0.733& 0.750& 0.958& 0.800& 0.733& 0.769& 0.820\\
    & 2020 & 0.786& 0.840& 0.857& 0.815& 0.769& 0.920& & 0.786& \\
recall\textsuperscript{${\dagger\dagger}$} & 2019 & 0.800& 0.680& 0.880& 0.720& 0.920& 0.800& 0.880& 0.800& 0.827\\
    & 2020 & 0.880& 0.840& 0.720& 0.880& 0.800& 0.920& & 0.880& \\
specificity & 2019 & 0.920& 1.000& 0.680& 0.760& 0.960& 0.800& 0.680& 0.760& 0.819\\
    & 2020 & 0.760& 0.840& 0.880& 0.800& 0.760& 0.920& & 0.760& \\
\hline
 & & \multicolumn{9}{c}{\textbf{GAT}}\\
\hline
CAR\textsuperscript{*} & 2019 & 0.860& 0.840& 0.760& 0.760& 0.940& 0.780& 0.820& 0.800& 0.825\\
    & 2020 & 0.780& 0.840& 0.780& 0.840& 0.800& 0.920& & 0.860& \\
F1\textsuperscript{$\dagger$} & 2019 & 0.851& 0.818& 0.786& 0.760& 0.939& 0.784& 0.824& 0.808& 0.825\\
    & 2020 & 0.776& 0.833& 0.756& 0.846& 0.808& 0.920& & 0.863& \\
precision\textsuperscript{$\ddagger$} & 2019 & 0.909& 0.947& 0.710& 0.760& 0.958& 0.769& 0.808& 0.778& 0.828\\
    & 2020 & 0.792& 0.870& 0.850& 0.815& 0.778& 0.920& & 0.846& \\
recall\textsuperscript{${\dagger\dagger}$} & 2019 & 0.800& 0.720& 0.880& 0.760& 0.920& 0.800& 0.840& 0.840& 0.821\\
    & 2020 & 0.760& 0.800& 0.680& 0.880& 0.840& 0.920& & 0.880& \\
specificity & 2019 & 0.920& 0.960& 0.640& 0.760& 0.960& 0.760& 0.800& 0.760& 0.829\\
    & 2020 & 0.800& 0.880& 0.880& 0.800& 0.760& 0.920& & 0.840& \\
\hline
\end{tabular}}
\resizebox{0.95\linewidth}{!}{
\begin{tabular}{l}
\textsuperscript{*} CAR = classification accuracy rate\\
\textsuperscript{$\dagger$} F1 score = $2(\text{recall}^{-1}+\text{precision}^{-1})^{-1}$.\;\mbox{\hspace{4in}}\;\\
\textsuperscript{$\ddagger$} precision = sensitivity; \\
\textsuperscript{${\dagger\dagger}$} recall = positive prediction value (PPV)\\
\textsuperscript{\#} For Dartmouth, the labeled messages  in the testing data are from 2019 as all labelled messages  \\
in 2020 are included  in the training data. \\
\hline
\end{tabular}}
\end{table}

The classification results can be summarized as follows. First, each examined metric reaches at least 80\% for all three classification approach. Second, compared to  GAT and RoBERTa, model-stacking further improve the overall classification accuracy, with higher values in almost all the examined metrics. Third, there is some variation in the prediction accuracy  across the  schools and years for the examined metrics, there is not a single school or in a single year that achieves the best performance in every metric, which is expected given the heterogeneity of the data. 

Compared to the results in the literature in the accuracy of sentiment classification using social media data via ML techniques, which typically ranges from 70\% to 85\%  \cite{wang2020covid, chakraborty2020sentiment, gupta2020sentiment, rahman2022exploring, imran2020cross}, our classification accuracy rates are among the best, if not the best. 

We also summarize the proportions of negative sentiment, overall, by year, and by school, respectively and the results are presented in Table \ref{tab:proportion}. Specifically, the percentage of negative sentiment is 39.14\% overall, 35.93\% in 2019, 41.75\% in 2020 (a 5.82\% increase in absolute proportion and 16.20\% increase in the proportion of negative sentiment from 2019). The school-specific purporting ranges from 24.18\% to 39.88\%  across the 8 schools in 2019 and from 23.16\% to 53.23\% in 2020. Among the schools, ND and UMich have the biggest jumps in the negative sentiment proportion from 2019 to 2020, which is an interesting observation as both schools have highly profitable football programs that were not open to the public in 2019. The proportions of the negative sentiments in our study, both overall and across the schools,  fall within the typical range  of reported negative sentiment proportions in the literature ($\sim15\%$ to $\sim50\%$, which varies by population and geographical locations such as countries; in addition, and sentiments can change rapidly due to reacting to announcements or implementations of new policies such as lock-down, social distancing, hygiene measures) and  are consistent with the trend of negative sentiments during the pandemic vs. pre-pandemic reported in the literature  \cite{bhat2020sentiment, gupta2020sentiment,  vijay2020sentiment, dubey2020twitter}.
\begin{table}[!htb]
\centering 
\caption{\bf Meta-model labelled sentiment frequency and percentage by school-year}\label{tab:proportion}
\resizebox{0.8\textwidth}{!}{
\begin{tabular}{lccccc} 
\hline
School &Year& \# Negative & \# non-Negative & Total \# & \%Negative\\
\hline
\multirow{2}{5em}{UCLA} & 2019 & 6121 & 10773 & 16894 & 36.23\\
 & 2020 & 7680 & 11034 & 18714 & 41.04\\
\hline
\multirow{2}{5em}{UCSD} & 2019 & 6502 & 11000 & 17502 & 37.15\\
 & 2020 & 6696 & 10650 & 17346 & 38.60\\
\hline
\multirow{2}{5em}{UCB} & 2019 & 6650 & 10023 & 16673 & 39.88\\
 & 2020 & 7699 & 10495 & 18194 & 42.32\\
\hline
\multirow{2}{5em}{UMich} & 2019 & 4126 & 8398 & 12524 & 32.94\\
 & 2020 & 6922 & 7957 & 14879 & 46.52\\
\hline
\multirow{2}{5em}{Harvard} & 2019 & 935 & 1958 & 2893 & 32.32\\
 & 2020 & 1138 & 2416 & 3554 & 32.02\\
\hline
\multirow{2}{5em}{Columbia} & 2019 & 1686 & 3345 & 5031 & 33.51\\
 & 2020 & 3423 & 5914 & 9337 & 36.66\\
\hline
\multirow{2}{5em}{Dartmouth} & 2019 & 308 & 966 & 1274 & 24.18\\
 & 2020 & 284 & 942 & 1226 & 23.16\\
\hline
\multirow{2}{5em}{ND} & 2019 & 395 & 1184 & 1579 & 25.02\\
 & 2020 & 4232 & 3718 & 7950 & 53.23\\
\hline
\multirow{2}{5em}{total} & 2019 & 26723 & 47647 & 74370 & 35.93\\
 & 2020 & 38074 & 53126 & 91200 & 41.75\\
\hline
\end{tabular}}
\end{table}

\subsection{Statistical inference for the pandemic effect on sentiment}
The GLMM results are presented in Table \ref{tab:glmm}. The odds of negative sentiments in year 2020 was 1.257 times the odds of Negative in year 2019, after adjusting for school type and location, with a $p$-value $<0.001$.  ``In-person learning'' in 2020 also affected sentiments in a statistically significant manner and the odds of negative sentiments increase by 48.3\% compared to ``online learning''.  Whether a school is located in a small city and whether a school is private do not influence the odds of negative sentiments in a statistically significant manner in either model, though being in a small city was associated with numerically  higher odds of negative sentiments and private schools were associated with lower odds of numerically negative sentiments relative to public schools.
\begin{table}[!htb]
\caption{\bf Estimated effects of the pandemic and in-person learning on the odds ratios of negative sentiment using GLMM} \label{tab:glmm}
\begin{center}
\resizebox{0.8\textwidth}{!}{\begin{tabular}{ lcc c lcc} 
\hline
\multicolumn{3}{c}{Effect of pandemic} & &\multicolumn{3}{c}{Effect of in-person learning in 2020}\\
\cline{1-3}\cline{5-7}
Factor & odds ratio & $p$-Value && Factor & odds ratio & $p$-Value \\
\cline{1-3}\cline{5-7}
\textbf{Year 2020} & \textbf{1.257} & $\mathbf{<0.001}$  && \textbf{In-Person} &  \textbf{1.483} & \textbf{0.029}\\
Small City &  1.076 & 0.619 && Small City  & 1.172 & 0.343\\
Private & 0.787 & 0.146 && Private &  0.765 & 0.130 \\
\hline
\end{tabular}}
\end{center}
\end{table}

To examine the sensitivity of the statistical inference to the predicted outcomes via ML approaches, we run the same GLMM models on the data with the predicted outcomes via  RoBERTa and GAT, respectively. The results are  shown in Appendix \ref{S1_File}. Overall, the results are similar to Table \ref{tab:glmm} (the directions of the effects of the factors and the statistical significance are consistent with those in Table \ref{tab:glmm}), implying robustness of the statistical inference to the choice of a ML labelling approach in this study, though  there are some numerical differences around the estimated odds ratios, as expected.

\section{Discussion}
\label{sec:discussion}
In this study, we collected social media data from Reddit and applied state-of-the-art ML techniques, together with statistical modelling and inference,  to study whether the pandemic has negatively affected the emotional and psychological states of a sub-population. Social media data are real-world data (RWD) and they have the typical RWD characteristics, such as observational,  unstructured (e.g., textual, relational), heterogeneous, voluminous, among others. Through pre-training via the NLP technique RoBERTa, the unstructured and textual data are transformed to real-valued vectors, with important textual and contextual information in the messages preserved. We also tried other word embedding approaches and the results are much worse. Relatively, the additional relational information in this study (binary directional edge) is less useful, compared to the rich semantics information extracted through the powerful NLP procedure. 

We controlled two factors during the data collections process -- time period and school. For the choice of time period, since the goal of the study is the pre-pandemic vs. pandemic comparison when school is in session, Aug to Dec in 2019 and 2020 are reasonable choices. In the future, we plan to keep collecting Reddit data from the same period every year (2021 and beyond) to examine the long-term effects of the pandemic on the emotional and psychological states and whether and when the states will return to the pre-pandemic baseline.  For the choice of schools, in this study, we focus on a limited set of the R1 doctoral universities per the Carnegie classification, mainly due to computational and data storage limitation; we plan to expand the study to include  schools from different categories in future studies. Among the selected subset of schools, we  balanced location, type of schools, online vs.~in-person learning  (for the last one, we made sure each school has clearly stated whether they adopted online or in-person leaning officially on their website).

This study serves as a good example of using  ML techniques and statistical modeling and inference in one application to make sense and better use of RWD. Specifically, rather than focusing on and stopping at just demonstrating the accuracy of a ML classification approach, we use the ML-labelled data to perform statistical inference and understand relationships between explanatory variables of interest and the outcome through GLMM, controlling for possible confounders. The latter goal could have been achieved through a controlled experiment  -- recruiting individuals, administering questionnaire to measure their sentiment  before the pandemic and during the pandemic, and then comparing the sentiment between the two time periods. However, this experiment would be much more costly in time and resources and there could be a large recall bias when the subjects tried to remember what their sentiments were a year ago before the pandemic as the study would not exist in the first place if it were not for the pandemic. 

The conclusions we drew from the GLMM analysis regarding the effects of the pandemic and in-person learning on the sentiment cannot be blindly generalized to the whole population; instead, the conclusions only pertain to the sub-population where the sample data in this study come from. Regarding what this sub-population is, it is impossible, at least from our end, to characterize its full demographic profile with just the social media (we do not have access to users' profile information due to privacy reasons). As a result, we may only draw a very sketchy profile for the sub-population that includes: those who were active Reddit users from Aug to Oct in 2019 and 2020 and who were related to the eight R1 schools in some way (e.g., current students, alumni, faculty, staff in those schools, their relatives or friends, people who live nearby or are interested in certain aspects of the school such as prospective students, sports fans, etc). On the other hand, despite the lack of the traditional specifics on what this sub-population is, the conclusions are a meaningful addition to the  growing real-world evidence and  help to complete the whole picture on the various types of impacts of the pandemic that are consistent across different sub-populations.

There is a couple of limitations in the study. First, as mentioned in Section \ref{sec:labeling}, we focused on binary classification in this study due to the relatively low accuracy in  classification with more granularity, which is a rather generic problem in sentiment analysis instead of being a particular problem to our study.  Working with more granular classifications on sentiment would likely help with more thorough studying of the impacts of the pandemic.  Second, we considered only one general node type ('message') and one relation type (`reply to') as the graph input (e.g. a homogeneous information graph) to the GAT NN model while the collected data have richer information on both node type (e.g., `message' can be sub-categorized into 'original post', 'post reply', 'comment reply'; 'user' can be another node type) and  relation type (besides 'reply to', a user may 'up-vote/down-vote', 'share', 'save', 'follow', or 'report' a post/comment) that the current study does not fully utilize. 
Training with heterogeneous information graphs may lead to more accurate classification results than the GAT results in this study.
Lastly,  the GLMM models assume the messages are correlated if they are from the same school, and but they do not consider the potential correlation among the messages from the same user.  On the other hand, given the large number of users and the wide range on the number of posted messages  across users (one to tens, either hundreds), this would lead to to highly unbalanced data and possible convergence issues during optimization and unstable results if the correlations among the users are modelled. We plan to conduct a follow-up study to collect more data in more schools and from after 2020 and aim to address each of the above limitations in the follow-up study.

\section{Conclusions}
\label{sec:conlusions}
We  developed a model stacking approach for learning sentiment labels from the real-world social media data. The components in the ensemble includes  RoBERTa and GAT that leverage the semantics/textural and graph/relational information, respectively, in the  data. The ensemble approach achieved greater than 80\% prediction accuracy on sentiment overall by various metrics. The subsequent statistical modelling and inference based on the labelled data suggest the pandemic has a negative impact on  the group's  emotional and psychological states in a statistically significant manner and in-person  teaching also increases the odds of negative sentiment in a statistically significant manner compared to online learning during the pandemic.  This study provides a good example of using both ML techniques and statistical modeling and inference to make sense and better use of the RWD and generating useful real-world evidence that might be of interest to different stakeholders. 

\section*{Appendix}
\paragraph*{A: Data and Codes.} \label{S1_File}

The data we collected from Reddit for this study include 1) userID,  2) time of a posted message, 3) content of a message, 4) dyadic relation between two messages in terms of whether one is a  reply to the other. There is no information in the collected data that allows direct identification of a user in the data.  The information in 'userID' was not used in our model training or methodological development and was replaced by dummy integers  after the data were downloaded.  

The data collection is in compliance with the Reddit privacy policy \cite{redditprivacy}, Reddit user agreement \cite{reddituser},  and Reddit API’s term of use \cite{redditapi}. The data are anonymized in the sense that we replaced the actual Reddit userIDs in the data with dummy IDs (integers  1, 2, ...), further limiting potential privacy risk (e.g., through record linkage) without losing information for learning task based on the data. The unprocessed data with actual Reddit userIDs replaced by dummy IDs are available at 
\url{https://drive.google.com/file/d/13PzwBAjyI4VYpCEVye6fU3gGsDvdnrBY/view?usp=sharing} and the RoBERTa-processed data and scores are available at 
\url{https://drive.google.com/file/d/1RRi5o1JCWeuS-dfhNOuxsk7JxJXKBOuL/view?usp=sharing}.

The code for this study can be download from \url{https://github.com/AlvaYan/Sentiment-Analysis-GNN-During-COVID19}). The  Python  code  for  the  RoBERTa framework that we applied  is  adapted  from \cite{barbieri2020tweeteval} and is available at \url{https://huggingface.co/cardiffnlp/twitter-roberta-base-sentiment};  the Python code for training the GAT NN is adapted from \cite{wang2019heterogeneous} (\url{https://github.com/Jhy1993/HAN}).

\paragraph*{B: 
Estimated effects of pandemic and in-person learning on the odds ratios of negative sentiment via GLMM based on the predicted outcome labels by RoBERTa and GAT.}
\begin{center}
\resizebox{0.87\textwidth}{!}{
\begin{tabular}{l| lcc |lcc} 
\hline
labelling & \multicolumn{3}{c|}{Effect of pandemic} & 
\multicolumn{3}{c}{Effect of in-person learning in 2020}\\
\cline{2-7}
method & Factor & odds ratio & $p$-Value &Factor & odds ratio & $p$-Value \\
\hline
& \textbf{Year 2020} & \textbf{1.310} & $\mathbf{<0.001}$  & \textbf{In-Person} &  \textbf{1.562} & \textbf{0.011}\\
GAT & Small City &  1.106 & 0.410 & Small City  & 1.231 & 0.229\\
& Private & 0.723 & 0.051 & Private &  0.686 & 0.052 \\
\hline
RoBERTa  &\textbf{Year 2020} & \textbf{1.196} & $\mathbf{<0.001}$  & \textbf{In-Person} &  \textbf{1.414} & \textbf{0.049}\\
& Small City &  1.073 & 0.641 & Small City  & 1.148 & 0.350\\
& Private & 0.839 & 0.219 & Private &  0.818 & 0.249 \\
\hline
\end{tabular}}
\end{center}

\section{Acknowledgments}
Tian Yan is a graduate student and Fang Liu is Professor in the Department of Applied and Computational Mathematics and Statistics at University of Notre Dame, Notre Dame (email correspondence: fang.liu.131@nd.edu). This work was supported by the China Scholarship Council and the University of Notre Dame Asia Research Collaboration Grant.



%
%
%

\end{document}